# Multi-AI Agent Framework Reveals the "Oxide Gatekeeper" in Aluminum Nanoparticle Oxidation

Yiming Lu[1,2], Tingyu Lu[1], Di Zhang[1], Lili Ye*[2], Hao Li*[1]

[1] Advanced Institute for Materials Research (WPI-AIMR), Tohoku University, Sendai 980-8577, Japan

[2] School of Chemical Engineering, Dalian University of Technology, Dalian, Liaoning 116024, China

* Corresponding author Email:

yell@dlut.edu.cn (L. Ye)

li.hao.b8@tohoku.ac.jp (H. Li)

## Abstract

Aluminum nanoparticles (ANPs) are among the most energy-dense solid fuels, yet the atomic mechanisms governing their transition from passivated particles to explosive reactants remain elusive. This stems from a fundamental computational bottleneck: *ab initio* methods offer quantum accuracy but are restricted to small spatiotemporal scales (< 500 atoms, picoseconds), while empirical force fields lack the reactive fidelity required for complex combustion environments. Herein, we bridge this gap by employing a **"human-in-the-loop" closed-loop framework where self-auditing AI Agents validate the evolution of a machine learning potential (MLP)**. By acting as scientific sentinels that visualize hidden model artifacts for human decision-making, this **collaborative cycle ensures quantum mechanical accuracy while exhibiting near-linear scalability to million-atom systems and accessing nanosecond timescales** (energy RMSE: 1.2 meV/atom, force RMSE: 0.126 eV/Å). Strikingly, our simulations reveal a temperature-regulated dual-mode oxidation mechanism: at moderate temperatures, the oxide shell acts as a dynamic "gatekeeper," regulating oxidation through a "breathing mode" of transient nanochannels; above a critical threshold, a "rupture mode" unleashes catastrophic shell failure and explosive combustion. **Importantly, we resolve a decades-old controversy by demonstrating that aluminum cation outward diffusion, rather than oxygen transport, dominates mass transfer across all temperature regimes, with diffusion coefficients consistently exceeding those of oxygen by 2-3 orders of magnitude**. These discoveries establish a unified atomic-scale framework for energetic nanomaterial design, enabling the precision engineering of ignition sensitivity and energy release rates through intelligent computational design.

# 1. Introduction

The quest for high-performance energy storage and propulsion systems has driven intense interest in aluminum nanoparticles (ANPs), which deliver exceptional energy densities of up to 31 kJ/g [1] and serve as critical components in advanced aerospace and defense technologies [2], [3], [4]. Despite decades of research, a fundamental mystery has persisted: how do these nanoscale powerhouses transition from benign, passivated particles at ambient conditions to explosive reactants at elevated temperatures? This dramatic behavioral shift spans an extraordinary range—from room-temperature oxide shell formation to violent high-temperature combustion involving complex gas-phase aluminum (Al) species. The underlying atomic-scale processes that control this transition have remained elusive, creating a critical knowledge gap that limits our ability to precisely engineer energy release rates and optimize combustion efficiency. Resolving this long-standing puzzle requires understanding the fundamental mass transport mechanisms that govern ANP reactivity across the complete temperature spectrum—a challenge that has defied conventional experimental and theoretical approaches due to the extreme spatiotemporal scales involved, from picosecond atomic motions to millisecond macroscopic energy release [8], [9], [10].

Accurate simulation of ANP combustion faces a fundamental scientific challenge: achieving large-scale spatiotemporal simulations while maintaining quantum mechanical accuracy. Current mainstream theoretical methods suffer from inherent limitations. Quantum mechanical approaches, exemplified by density functional theory (DFT), can precisely describe interatomic interactions and electronic structures but are computationally prohibitive, restricting simulations to hundreds of atoms and picosecond timescales—far from covering complete combustion evolution [11], [12]. Conversely, classical reactive force fields like reactive force field (ReaxFF) can handle larger scales (thousands of atoms, nanosecond timescales) [7], [13] but rely on empirical parameter fitting. This limitation often confines studies to phenomenological observations rather than establishing rigorous physical laws. A critical question arises: is the previously observed dominance of Al outward diffusion a genuine physical phenomenon, or merely an artifact of parameterization? To transit from qualitative simulation to quantitative confirmation, a method combining quantum precision with macroscopic scalability is required. Furthermore, ReaxFF often struggles to accurately capture the electronic structure rearrangements and charge transfer effects during complex physicochemical evolution [14]. Real ANP combustion processes involve microsecond to millisecond timescales accompanied by complex hierarchical nanostructure evolution and interfacial dynamics—all beyond the capability boundaries of existing computational methods.

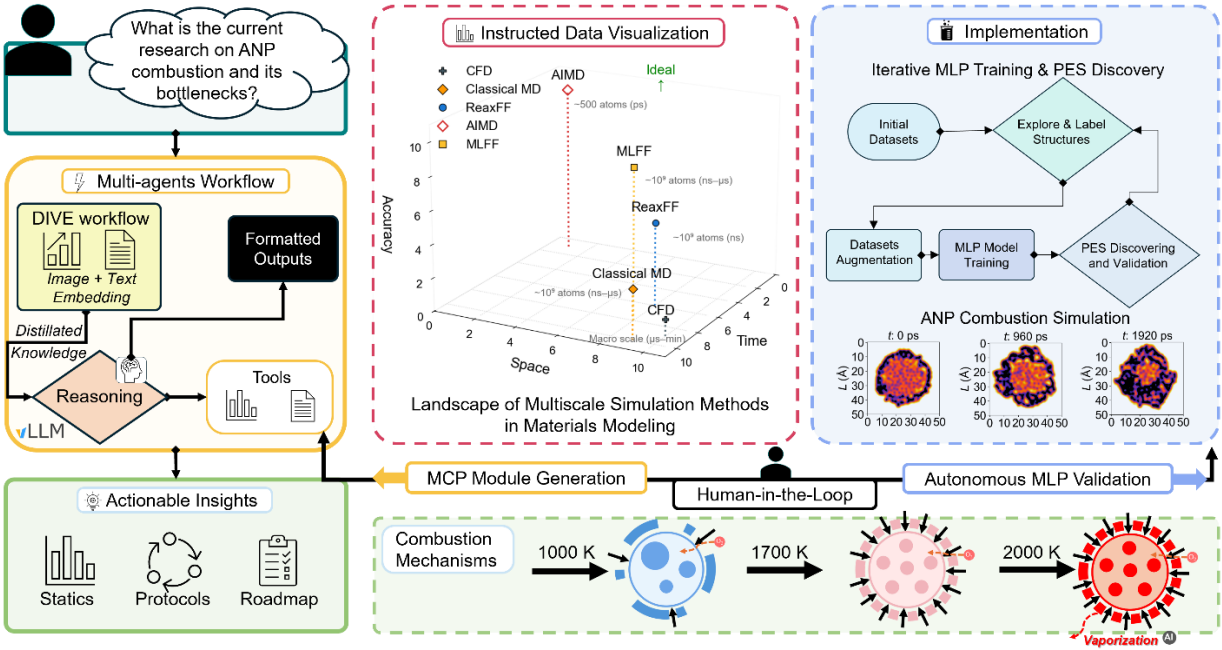

**Fig. 1 Integrated AI-assisted workflow for ANP combustion research.** The integrated framework combines our DIVE (Descriptive Interpretation of Visual Expression) system for automated literature analysis with machine learning potential (MLP) development, featuring an automated deployment protocol that continuously integrates trained MLPs to enhance the system's predictive and analytical capabilities for Al-O systems. This closed-loop architecture enables a faster and easier MLP development for the simulations of ANP oxidation mechanisms at unprecedented spatiotemporal scales (detailed technology implementation is provided in Fig. S7).

To address these fundamental challenges, we developed an advanced self-reinforcing multi-AI Agent framework (Fig. 1) that autonomously navigates the research landscape. The Agent systematically evaluates existing methodologies through intelligent literature analysis, identifies machine learning potential (MLP) as the transformative solution to the long-standing "accuracy-efficiency" dilemma [15], [16], [17], and executes a rigorous development pipeline. The approach learns complex potential energy surfaces from DFT data, achieving near-first-principles accuracy with the computational efficiency of classical force fields. By employing an active learning strategy guided by the AI Agent's evolving understanding, the required amount of DFT training data can be substantially reduced while maintaining predictive reliability [18], [19]. For ANP combustion simulation, a robust MLP training framework requires three decisive advantages: **1)**

**High fidelity:** ensuring exceptional predictive accuracy through direct learning from DFT data rather than empirical parameter fitting; **2) High efficiency:** computational speed improved by several orders of magnitude compared to DFT, enabling microsecond-scale long-duration simulations; **3) High compatibility:** smooth integration with mainstream molecular dynamics software like LAMMPS [20]. Among available MLP implementations, the DeepMD-kit package, particularly its attention-based DPA-2 architecture, emerges as the optimal choice. It offers superior representational capacity for complex chemical environments while maintaining high parallel efficiency and compatibility with our self-reinforcing AI framework [21], [22].

Herein, we employ this AI-driven workflow to develop a high-fidelity MLP for the Al-O system *via* active learning with enhanced sampling, achieving quantum mechanical accuracy at previously unattainable scales. Distinguishing our approach from traditional linear workflows, we construct a closed-loop system where a **"self-auditing" AI Agent acts as a necessary bridge between training and application**. Instead of blindly deploying trained models, the Agent employs the Module Context Protocol (MCP) to conduct rigorous "scientific audits"—**quantitatively assessing model fidelity and visualizing error distributions**. This creates a decision checkpoint: the Agent presents evidence, and **the human expert decides whether to close the loop** by deploying the model or refining the training set, ensuring that only high-precision potentials drive the simulation. Leveraging this robust framework, our atomistic simulations reveal a temperature-regulated dual-mode oxidation mechanism: oxide-shelled ANPs transition from diffusion-limited oxidation at moderate temperatures to direct shell rupture at elevated temperatures. **Importantly, we resolve a decades-old controversy by demonstrating that Al diffusion, not O transport, dominates mass transfer in both bare and shelled nanoparticles across all temperature regimes**. These AI-guided discoveries establish a complete, atomic-level framework for ANP oxidation, paving the way for the precision engineering of energetic nanomaterials.

## 2. Results and Discussion

2.1 Machine Learning Potential Validation and Dataset Coverage

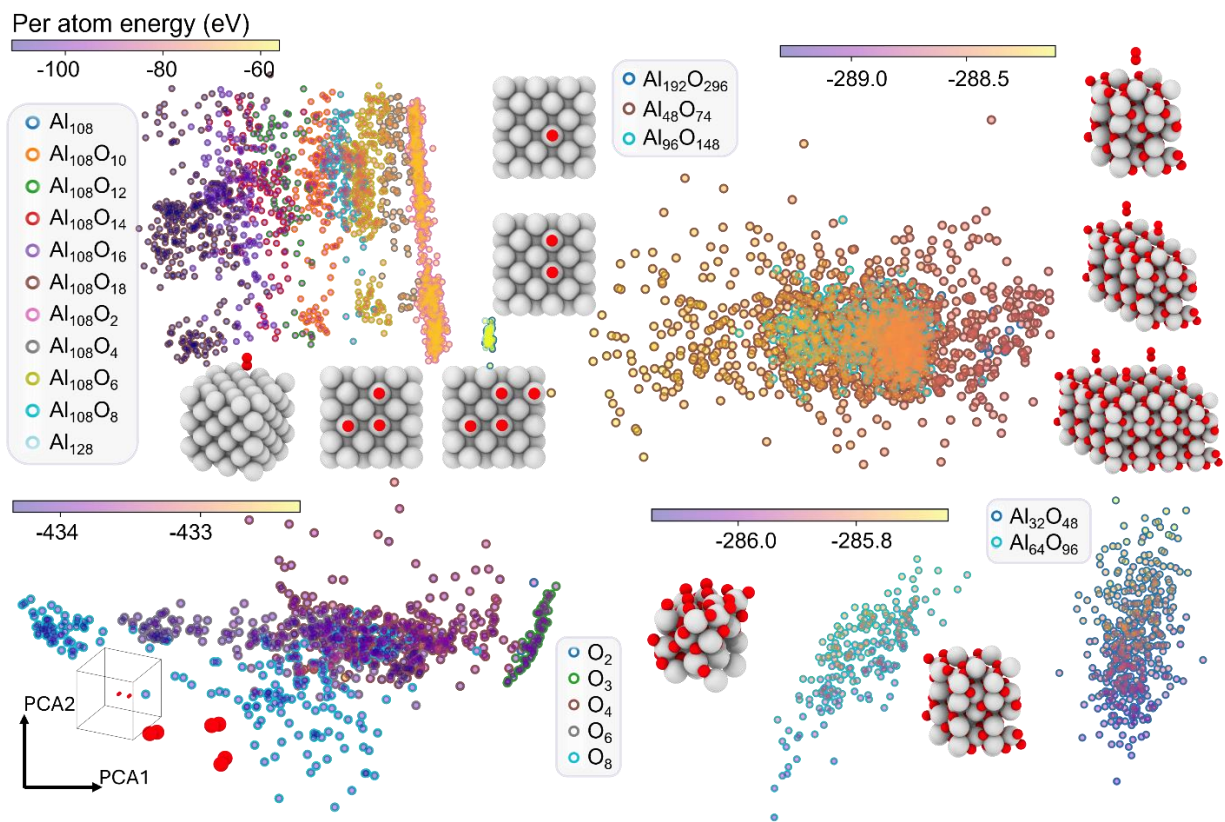

**Fig. 2 Atomic configuration space mapping and active learning validation based on smooth overlap of atomic positions (SOAP).** The figure shows the results of dimensionality reduction visualization of 88,833 training configurations through principal component analysis (PCA). Each data point represents an independent structure, with its position in the two-dimensional PCA space determined by the first two principal components of SOAP descriptors. The color scale represents per-atom energy, from blue (low-energy) to red (high-energy).

With the assistance of our multi-AI Agent framework (Fig. 1), accessible *via* a dedicated frontend interface (see **Supplementary Video 1** for a demonstration), we employed an iterative active learning strategy incorporating time-stamped Force-bias Monte Carlo (tfMC) sampling (detailed in **Methods** and **Supplementary Information**) to efficiently explore the Al-O reaction space. After approximately 80 iterations, we obtained a high-quality dataset of 88,833 configurations covering temperatures from 200 to 3000 K. Fig. 2 demonstrates the comprehensiveness of our high-quality training dataset within the chemical configuration space. Through principal component analysis (PCA) dimensionality reduction of smooth overlap of atomic positions (SOAP) descriptors [23], [24], we reveal clear structure-energy relationships where configurations with similar chemical environments form distinct clusters. Systems of different stoichiometric ratios occupy different regions, with low-energy equilibrium configurations concentrated at cluster centers and high-energy transition states distributed at cluster edges. This distribution confirms that our active learning strategy successfully captured rare events crucial for ANP combustion simulation, including oxygen diffusion pathways, Al ion migration, and phase transition environments.

Building on this comprehensive dataset, systematic comparison with classical force fields reveals the superior performance of our MLP model (Fig. S2). The model achieves energy root mean square error (RMSE) of 1.2 meV/atom and force RMSE of 0.126 eV/Å on testing data, representing **1-2 orders of magnitude improvement over traditional potentials**. Crucially, this validation was not merely a retrospective check but an active checkpoint facilitated by the AI Agent. By invoking the MCP tool to generate real-time error histograms and force parity plots, the Agent allowed us to inspect the error distribution tails explicitly, ensuring that no "black box" outliers compromised the simulation's physical validity. This Agent-assisted audit closed the verification loop, providing the strictly required evidence for the human expert to authorize the transition from MLP training to large-scale production simulations. Even for complex non-stoichiometric oxide structures, the MLP maintains high accuracy while traditional potentials show dramatically increased errors. The scatter plots demonstrate that MLP predictions are tightly distributed around the ideal diagonal, with over 99% of force errors concentrated within < 0.1 eV/Å. This exceptional accuracy across diverse chemical environments ensures reliable force fields for complex multiscale ANP combustion processes, providing unified description of metallic, covalent, and ionic interactions. Detailed validation results including structural benchmarks, thermodynamic properties, and phonon calculations are provided in the **Supplementary Information** (Figs. S3–S5). Crucially, the MLP accurately reproduces the phonon dispersion relations and thermodynamic properties (heat capacity and entropy) of both Al and $Al_2O_3$ up to 1000 K. As heat capacity determines the temperature rise rate and entropy correlates with phase transition latent heat, this thermodynamic fidelity, often lacking in classical potentials, is essential for the reliable prediction of combustion dynamics.

## 2.2 Temperature-Dependent ANP Oxidation

Leveraging the validated MLP model, our large-scale molecular dynamics simulations establish a unified framework for ANP oxidation that resolves long-standing debates over its governing mechanisms. We reveal a temperature-controlled, dual-mode mechanism where the nanoparticle's reactivity is ultimately dictated by the structural integrity of its passivating oxide shell. This shell acts as a dynamic gatekeeper, modulating the intrinsically dominant outward migration of Al atoms, which we identify as the fundamental rate-controlling process across all regimes. Additionally, we provide visualizations of these dynamic atomistic processes in the Supplementary Videos.

### 2.2.1 Intrinsic Transport Mechanisms in Bare Metal Nanoparticles

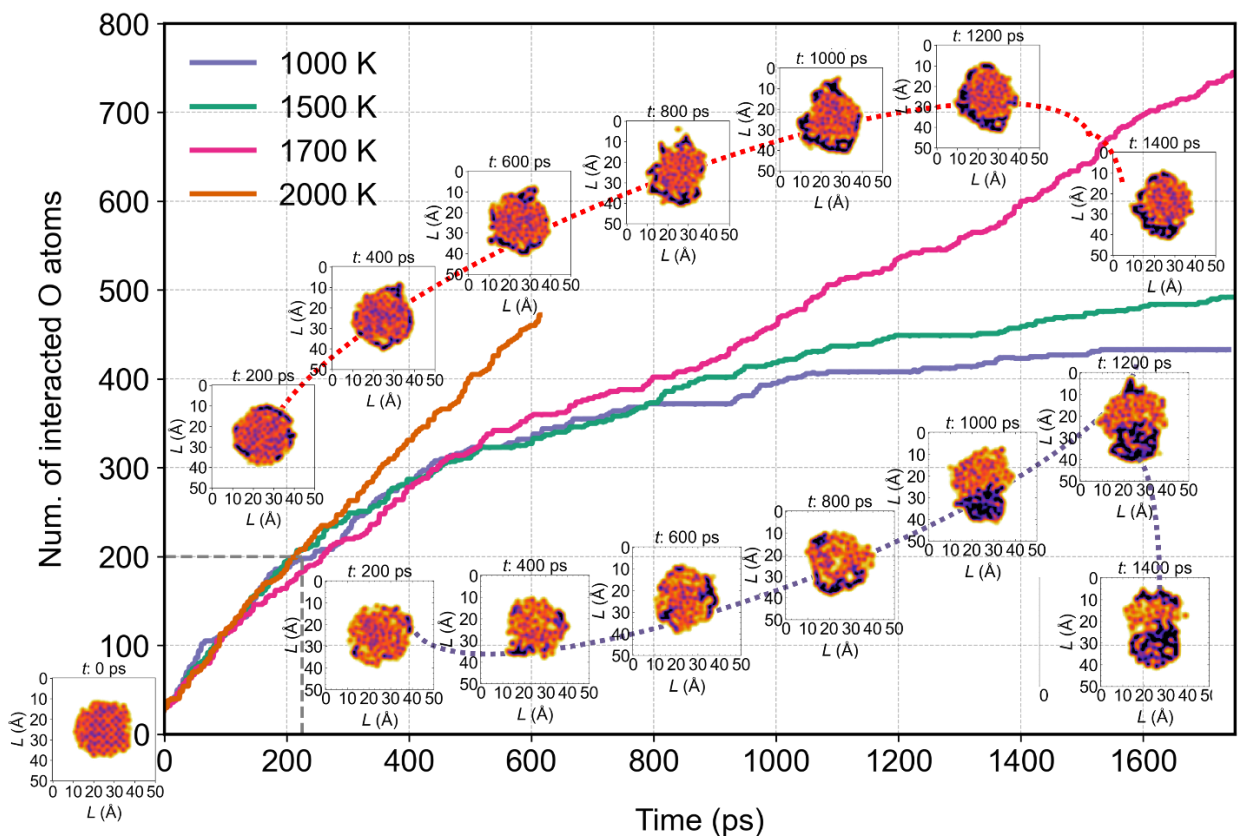

**Fig. 3 Two-stage oxidation kinetics in bare metal ANP.** Initial oxygen-supply limited phase (0-200 ps) transitions to temperature-dependent diffusion control, with 1000 K showing surface passivation while higher temperatures sustain more rapid oxidation.

To uncover the intrinsic oxidation mechanism of ANP, we first investigated bare metal ANPs, isolating the reaction from the complexities of a pre-existing oxide shell. Our simulations reveal a universal, two-stage process. Initially, oxidation is limited by the supply of oxygen, with the reaction rate being nearly identical across all temperatures (1000-2000 K) for the first 200 ps (Fig. 3). This temperature independent behavior confirms that the bare ANP surface is highly reactive, readily consuming all available oxygen and establishing a consistent oxidized state before temperature-specific mechanisms begin to dominate. Following this brief initiation phase, the oxidation pathway diverges into three distinct, temperature-controlled regimes, each characterized by progressively enhanced Al mobility and fundamentally governed by the superior transport properties of Al atoms compared to O. This finding aligns with experimental observations [25] showing that oxygen inward diffusion dominates oxidation at low temperatures (< 600 °C), while Al outward diffusion becomes the primary driving force at high temperatures (> 600 °C).

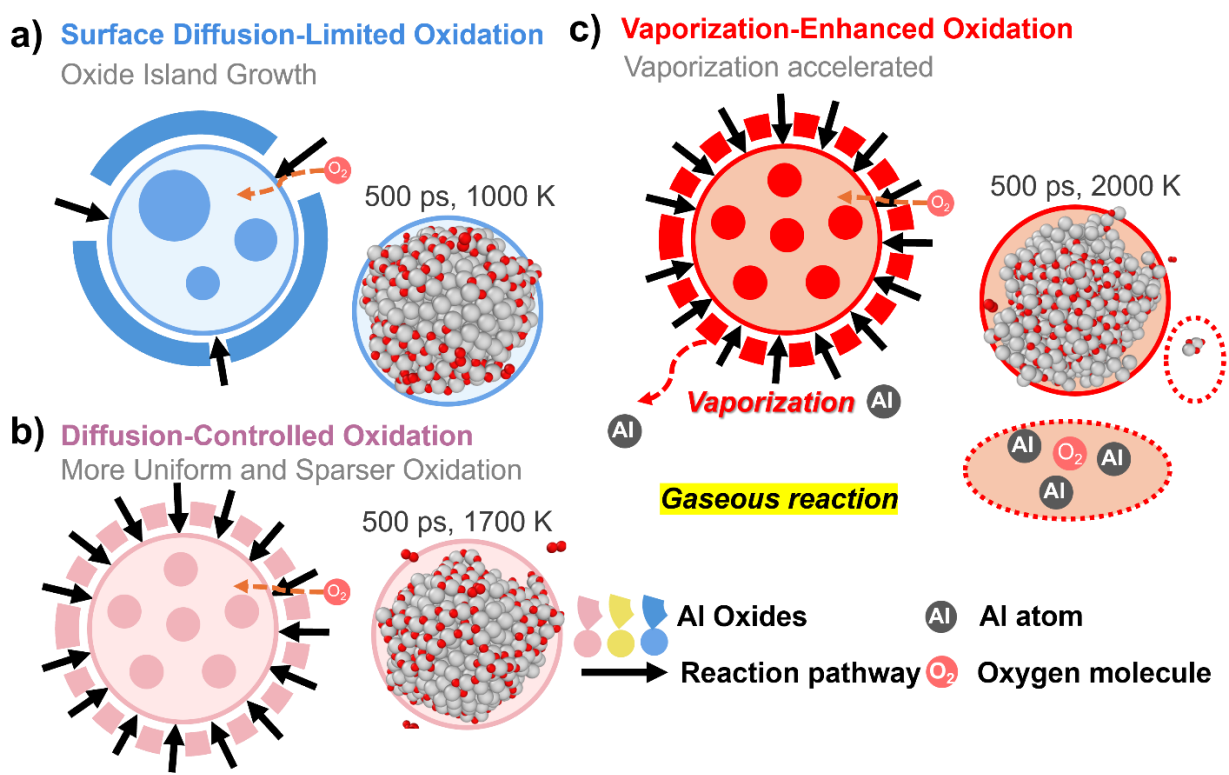

**Fig. 4 Schematic of temperature-dependent oxidation mechanisms for bare metal ANPs.** (a) At low temperatures, oxidation is limited by surface diffusion. O atoms have low mobility on the Al surface, leading to the formation of discrete, amorphous oxide islands rather than a uniform shell. (b) At intermediate temperatures, enhanced atomic mobility allows for a more uniform and complete, albeit sparser, oxide layer to form, representing a diffusion-controlled oxidation regime. (c) At high temperatures, oxidation is significantly accelerated by the vaporization of Al atoms from the nanoparticle surface. These vaporized Al atoms react with $O_2$ in the gas phase, leading to rapid and complete combustion.

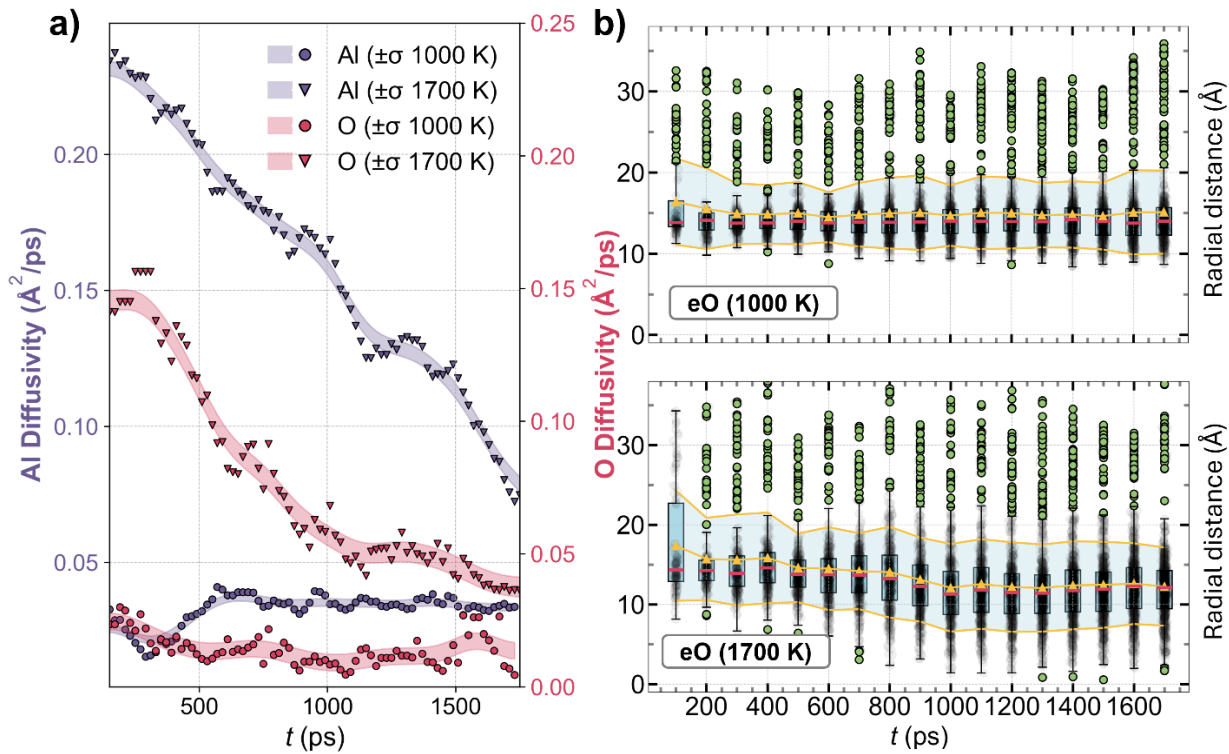

**Fig. 5 Al-dominated transport mechanisms.** Diffusion coefficients at 1000 and 1700 K demonstrate Al's consistently higher mobility than oxygen, confirming Al migration as the rate-controlling process. Oxygen inward migration during oxide layer growth. Radial distribution evolution demonstrates oxygen penetration from surface adsorption sites toward the particle core, forming the growing oxide shell.

The first temperature regime occurs at low temperature (1000 K), where oxidation proceeds through a thermodynamically-controlled **island nucleation and growth mechanism**, fundamentally dictated by the kinetic asymmetry between Al and O surface transport. The initial dissociative chemisorption of $O_2$ molecules establishes discrete amorphous oxide nuclei (Fig. 4a), which subsequently evolves according to classical heterogeneous nucleation theory. The critical mechanistic insight emerges from the pronounced difference in activation barriers: oxygen adatoms experience severe kinetic trapping with surface migration barriers of 1.25 eV (quantified via PLUMED OPES [26] on Al(100) facets, Fig. S6), rendering them effectively immobile at the available thermal energy ($k_BT \approx 0.086$ eV). Conversely, Al atoms maintain substantial surface mobility (Fig. 5), while O atoms struggle to diffuse inward as indicated by the barely changed radial distance distribution, creating a dynamic substrate beneath the static oxide patches. This kinetic disparity precludes uniform film formation and instead drives a preferential edge-growth mechanism, where the high chemical potential gradient at the three-phase boundary (oxide-metal-gas) promotes selective $O_2$ adsorption and reaction at island perimeters. The resulting three-dimensional island expansion follows Volmer-Weber growth kinetics until coalescence events progressively reduce the reactive surface area as indicated in Fig. 3. This behavior differs from the chain-like oxide nucleation reported by Zhang et al. using ReaxFF [27], yet both studies corroborate a nucleation-and-growth paradigm in ANP oxidation.

As temperature increases to the intermediate regime (1700 K), the oxidation mechanism undergoes a fundamental transition to **uniform oxide shell growth** (Fig. 4b), driven by the thermally-activated reconstruction dynamics of the molten ANP core. The elevated thermal energy ($k_BT \approx 0.146$ eV) drives the continuous surface restructuring, effectively disrupting the kinetic stabilization of discrete oxide islands observed at lower temperatures. This dynamic surface reconstruction creates a homogeneous reaction interface that prevents the formation of stable oxide nuclei and overcomes the previously limiting oxygen adatom mobility constraints. The enhanced thermal activation facilitates the predominant migration of Al atoms (Figs. 3 and 5) through a Wagner-type oxidation mechanism, where the chemical potential gradient drives Al diffusion to the reaction front. This Al-dominated mass transport enables the formation of a spatially uniform, continuously growing oxide layer that maintains structural coherence while accommodating volume expansion. The resulting diffusion-controlled oxidation kinetics follow parabolic growth laws, establishing a sustained reaction regime governed by the intrinsic mobility advantage of Al over O in the oxide matrix.

At the higher temperatures in the extreme regime (> 1700 K), the oxidation mechanism transcends conventional solid-state diffusion limitations through the **emergence of direct Al vaporization as an alternative kinetic pathway, which becomes dominant under higher temperature as verified in experimental research** [28]. At these elevated thermal conditions ($k_BT \geq 0.146$ eV), Al atoms acquire more thermal energy to overcome surface binding energies and undergo desorption-mediated mass transport (Fig. 4c). This gas-phase reaction pathway fundamentally alters the reaction kinetics by eliminating the rate-limiting step of solid-state diffusion through the oxide matrix. Quantitative analysis reveals a sharp thermodynamic threshold: while Al atom consumption remains negligible in the moderate temperature range (1000-1700 K), the desorption flux exhibits exponential temperature dependence, with approximately 40 Al atoms consumed within 1500 ps at 2000 K (Fig. S11). The enhanced reactivity manifests through dramatically increased O₂ consumption rates, where the rapid Al vaporization creates a highly reactive gas-phase environment that efficiently utilizes the periodically replenished oxygen supply. This vaporization-controlled mechanism bypasses the inherent transport limitations of condensed-phase diffusion, establishing a direct gas-phase reaction pathway that accounts for the explosive nature of ANP combustion at high temperatures. The transition represents a fundamental shift from transport-limited to thermodynamically-driven kinetics, where the reaction rate becomes governed by the Al vapor pressure rather than solid-state diffusion coefficients.

Our results on bare nanoparticles reveal that Al-dominated transport is the universal engine of oxidation. However, the reaction's macroscopic behavior is dictated by temperature, which modulates the mechanism from self-limiting island formation at low temperatures, to steady shell growth at intermediate temperatures, and finally to explosive, vaporization-driven combustion at high temperatures.

2.2.2 Oxide Shell Gatekeeper Effects

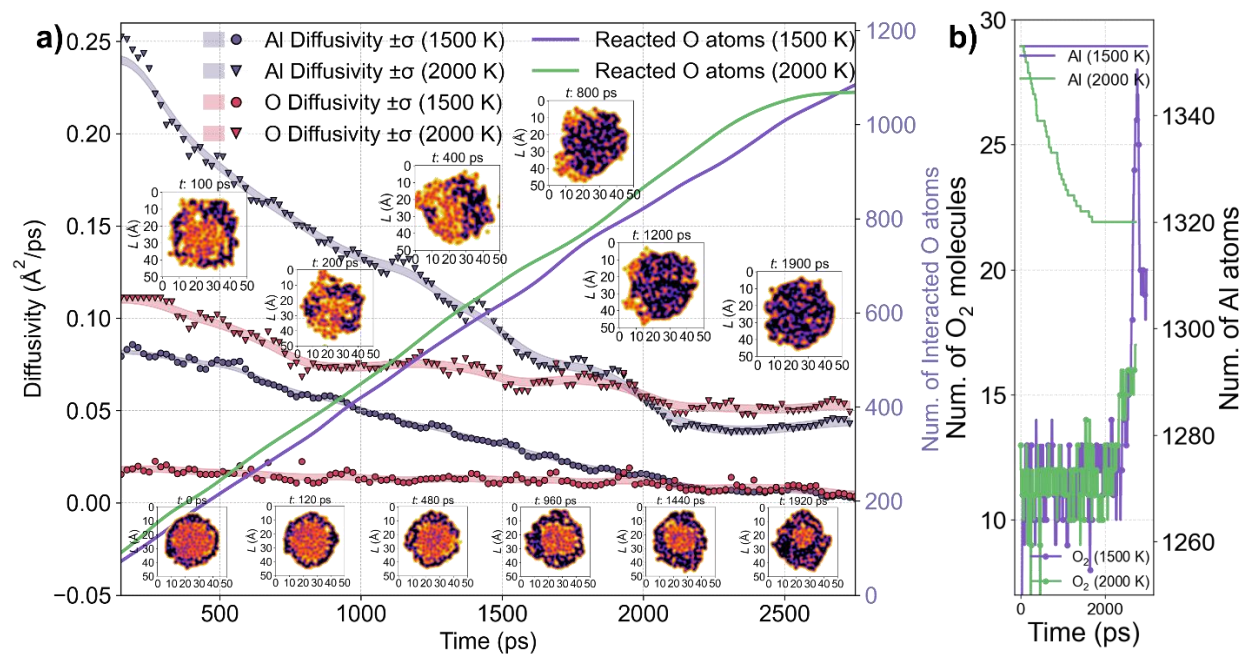

**Fig. 6 Oxide shell gatekeeper mechanisms modulating Al transport**. (a) Cross-sectional Gaussian density distributions reveal two distinct modes: "Breathing mode" at 1500 K with dynamic structural fluctuations creating transient transport channels, and "Rupture mode" at 2000 K with catastrophic shell failure exposing the reactive core. Diffusion coefficients consistently show Al mobility exceeding oxygen transport, confirming Al-dominated mechanisms persist even with shell constraints (detailed cross-sectional views at different temperatures are shown in Figs. S12 and S13). (b) Temporal variation of O₂ molecules and Al atoms at different temperatures.

Building on the intrinsic Al-dominated transport mechanism established in bare nanoparticles, we demonstrate that the presence of a pre-existing oxide shell introduces a sophisticated **"gatekeeper function"** that modulates access to reactive sites without altering the fundamental transport physics. This thermodynamically-responsive gatekeeper operates through two distinct, temperature-dependent kinetic regimes that represent different manifestations of the same underlying Al\ transport mechanism under progressively enhanced thermal conditions.

**Breathing Mode:** In the moderate temperature regime (1500 K), the oxide shell exhibits thermodynamically-controlled structural dynamics characterized by a **"breathing mode"** mechanism. As demonstrated in Fig. 6a, this regime is governed by the interplay between thermal activation energy and shell cohesive forces, resulting in reversible structural fluctuations that generate transient nanoscale transport pathways. Unlike rigid lattice diffusion, these dynamically-formed channels function as kinetic bottlenecks [25], establishing a diffusion-limited regime where mass transport across the oxide-metal interface becomes the rate-determining step. The radial distances distribution analysis of different atom groups (Fig. S14) quantitatively confirms the asymmetric transport kinetics: while inward oxygen diffusion proceeds through thermally-activated hopping mechanisms with limited penetration depth, outward Al migration continues to dominate but now operates under constrained flux conditions due to the shell's selective permeability. This temperature-controlled process, mediated by defect-assisted transport through the dynamically "breathing" oxide matrix, establishes a quasi-steady-state oxidation regime characterized by sustained, diffusion-limited kinetics where the shell acts as a molecular sieve. These computational predictions are strongly **supported by *in situ* TEM observations** from Zhou et al. [29], where the emergence of surface protrusions and concavities at elevated temperatures (973-1208 K) provides direct experimental evidence of these localized breathing modes and transient transport channels.

**Rupture Mode:** When the temperature exceeds the critical thermodynamic threshold (2000 K), the system undergoes a catastrophic phase transition to a **"shell rupture mode"** governed by thermomechanical instability. At this temperature regime, the accumulated thermal stress exceeds the oxide shell's fracture toughness, triggering rapid structural failure within nanosecond timescales and exposing the highly reactive molten Al core to direct oxidant contact (Fig. 6a). This critical transition fundamentally redefines the reaction's rate-limiting mechanism, shifting from diffusion-controlled transport within the intact shell to external mass transfer limitations. The shell rupture event dramatically amplifies the system's reactive surface area and eliminates the primary kinetic barrier, thereby unleashing the intrinsic Al transport potential and enabling direct desorption and gas-phase vaporization (Fig. 6b). This mechanistic transition from transport-limited to thermodynamically-driven kinetics explains the observed shift to explosive combustion behavior. The experimental counterpart to this catastrophic failure is captured [29], where distinct shell-breaking phenomena and core exposure become prevalent at extreme temperatures (1373 K), mirroring our simulation results.

Importantly, across both regimes, our analysis consistently shows that the diffusion rate of Al atoms is significantly higher than that of oxygen atoms. This confirms that even in the presence of an oxide shell, outward Al migration remains the dominant mass transport mechanism. Temperature does not change this fundamental fact; instead, it controls the permeability of the oxide shell, which in turn dictates the overall rate of the Al-fed reaction. This critical temperature for rupture is also inversely related to shell thickness [30], [31], suggesting a pathway to tune nanoparticle reactivity by engineering the morphology of the ANP.

2.2.3 Unified Temperature-Modulated Transport Framework

Synthesizing our findings from both bare and core-shell nanoparticles, we arrive at a unified mechanism that reconciles the diverse reactive behaviors of ANPs through a single governing principle: **temperature-modulated Al transport controlling reaction site accessibility**. This framework confirms that the outwardly diffusing Al flux is the universal engine of oxidation, while the oxide shell acts as a thermodynamic valve. This mechanism operates through the intrinsically superior mobility of Al atoms relative to oxygen, with temperature serving as the critical parameter that determines the structural integrity and permeability of the passivating oxide shell, thereby controlling the number and uniformity of exposed reaction sites available for oxidation.

The mechanism manifests through two distinct kinetic regimes: **(1) Diffusion-limited regime**, where the intact oxide shell operates in "breathing mode," creating transient nanochannels that constrain Al flux and maintain controlled oxidation; and **(2) Transport-unlimited regime**, where thermomechanical shell rupture eliminates diffusion barriers, enabling direct Al vaporization and explosive combustion. This transition represents a continuous evolution of the same fundamental transport process rather than distinct mechanistic pathways.

This unified framework establishes a fundamental design principle for energetic nanomaterials: **reactivity control through structural engineering of the oxide shell**. The temperature-dependent shell behavior functions as a thermodynamically-responsive switch that modulates reaction kinetics by controlling the accessibility of reactive Al sites. This mechanism provides a theoretical foundation for tailoring ANP performance through morphological design parameters (shell thickness, core-shell ratio) and environmental conditions (ambient oxygen concentration, heating rate), enabling predictive control over the transition from controlled energy release to explosive combustion. The universality of Al-dominated transport across all temperature regimes suggests that this principle may extend to other metal-oxide nanoparticle systems, offering broader implications for the rational design of reactive nanomaterials.

## 3. Conclusion

In summary, guided by the multi-AI Agent framework, we have developed a high-fidelity MLP model using a specially designed active learning strategy to establish a unified atomic-scale framework for ANP combustion. This approach resolves the long-standing debate over mass transport mechanisms and reveals a temperature-controlled dual-mode reactivity mechanism. The MLP was instrumental in enabling large-scale simulations with quantum accuracy, allowing us to demonstrate conclusively that outward Al diffusion is the dominant mass transport process across different temperature regimes, from low-temperature passivation to high-temperature combustion. This finding deepens our understanding of ANP oxidation at the nanoscale, providing a new theoretical foundation for a field previously guided by conflicting theories.

Our simulations reveal that the reactivity of ANPs is governed by a delicate interplay between the intrinsic Al-dominated transport and the temperature-dependent structural integrity of the oxide shell. We have **identified two distinct operational modes:** a low-temperature "breathing" mode, where controlled, diffusion-limited oxidation proceeds through transient nanochannels, and a high-temperature "shell rupture" mode, which triggers explosive energy release. This dual-mode mechanism, coupled with our discovery of a transition from oxide island growth to direct evaporation in bare nanoparticles, provides a comprehensive roadmap for tuning the energy release characteristics of ANPs. These insights have immediate implications for the rational design of next-generation materials, enabling the precise control of ignition sensitivity and energy release rates through the targeted manipulation of oxide shell properties.

Looking forward, we envision a future where AI-driven, quantum-accurate simulations guide the *in silico* discovery of novel materials for the grand challenges of energy and propulsion. This approach, which combines MLP with physics-based simulations, offers a powerful tool for accelerating the design of complex reactive systems. The self-evolving AI Agent framework introduced here demonstrates the potential for autonomous scientific discovery through continuous self-improvement and automatic deployment of trained MLPs. This closed-loop architecture suggests that future materials discovery could transition from assistive tools to collaborative research partners capable of hypothesis generation. We hope this work encourages the community to further explore these methods, paving the way for a new era of accelerated, data-driven discovery.

## 4. Methods

### 4.1 Foundational Dataset Sampling

Accurate MLP require comprehensive sampling of the chemical reaction space to effectively represent the diverse local atomic environments that different elements can experience during reactions. For ANP combustion, the training dataset must capture diverse atomic configurations across the reaction potential energy surface, including equilibrium structures of various nanoparticle morphologies, reaction transition states for Al-O bond formation and breaking, interfacial configurations between Al cores and alumina shells, and defect structures that serve as reaction hotspots. Achieving such comprehensive coverage presents a fundamental challenge: traditional ab initio molecular dynamics (AIMD) provides quantum-accurate trajectories but at prohibitive computational cost, severely limiting accessible timescales and configurational diversity, while introducing too much data into the training set. To address this limitation, we adopt a two-stage model training strategy combining initial AIMD sampling with subsequent active learning expansion.

We begin by establishing a foundational dataset through AIMD simulations of bulk Al and $Al_2O_3$ at multiple temperatures using the canonical ensemble with 0.5 fs time steps for 4000 steps. While this conventional approach cannot adequately sample the vast configurational space required for combustion simulations—particularly the rare events and high-energy transition states that govern reaction kinetics—it provides essential equilibrium structures and thermal fluctuations that serve as a robust foundation for our active learning strategy. This initial dataset becomes the crucial starting point upon which the active learning algorithm builds to achieve comprehensive coverage of the reaction potential energy surface.

## 4.2 Active Learning Implementation

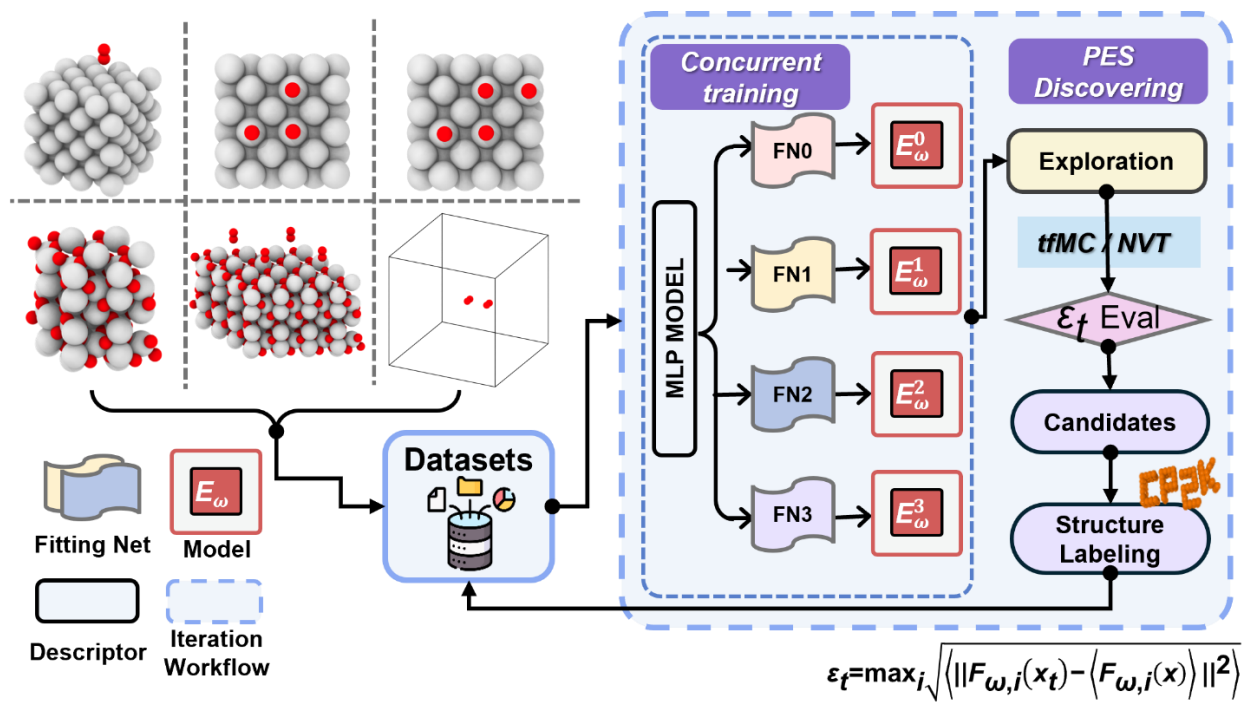

**Fig. 7 Flowchart of the active learning strategy employed in this study.** This cycle enhances the MLP by identifying informative atomic configurations through a Query-by-Committee (QBC) approach. Initially, four concurrent fitting networks (FN0–FN3) with identical hyperparameters but initialized with different random seeds are trained on the existing datasets. During the potential energy surface (PES) discovering phase, the configurational space is explored using tfMC/NVT MD. The model evaluates candidate structures by calculating the force deviation ($\varepsilon_t$); structures with uncertainty exceeding a predefined threshold are labeled via CP2K-based DFT calculations and incorporated into the training set. This process repeats until the MLP achieves the desired accuracy across the entire reaction space.

Building upon the foundational dataset established above, our active learning implementation systematically identifies and incorporates the most informative atomic configurations through an iterative approach that progressively enhances model accuracy by focusing computational resources on regions of high uncertainty (Fig. 7). The workflow operates through three integrated stages: ensemble model training, guided exploration using molecular dynamics coupled with tfMC simulation to overcome energy barriers, and uncertainty-based configuration selection for DFT labeling with CP2K [32]. This creates a self-improving cycle that systematically expands the model's predictive scope while ensuring comprehensive yet computationally efficient sampling of the potential energy surface.

To prevent overfitting of MLPs on specific regions of the potential energy surface from redundant configurations identified during active learning, we implement a GPU-accelerated structural similarity-based optimization strategy [19] using SOAP descriptors (Fig. S2). This approach, employing a furthest point sampling algorithm, systematically eliminates structurally similar configurations while preserving the most representative ones that capture essential chemical diversity. By leveraging GPU acceleration for descriptor calculation and clustering, we achieved over 100-fold speed improvements compared to CPU implementations. Detailed parameters for the MLP architecture (Table S1), training protocols, DFT calculations (Table S2), and datasets filtering details are provided in the **Supplementary Information**.

## 4.3 Self-Reinforcing Multi-Agent Framework

Our AI Agent framework transforms scientific literature into actionable computational insights through an integrated four-stage closed-loop pipeline (Fig. S7). The framework begins with systematic knowledge ingestion and vectorization using our DIVE (Descriptive Interpretation of Visual Expression) methodology, achieving 10%-15% performance improvement over commercial models in data extraction accuracy [33]. This extracted knowledge undergoes agentic analysis through large language model powered reasoning that systematically evaluates computational methods and identifies MLPs as the optimal solution for ANP combustion simulation. The framework then translates these analytical insights into concrete MLP development through automated insight generation and implementation protocols. The distinguishing feature is the establishment of a "Human-Agent-Model" closed feedback loop. In this loop, the researcher can use the Agent to autonomously invoke the validation suite to stress-test the MLP, generating visual analytics specifically designed to expose physical inconsistencies (e.g., atomic force inconsistencies). This output serves as the "decision support" signal for the human researcher, who then closes the loop by either approving the model for the subsequent simulation phase or injecting new physics-based constraints for retraining. This mechanism transforms the verified MLP from a black-box approximation into a thoroughly audited scientific instrument. The integrated AI Agent provides comprehensive computational functionalities including structure energy prediction, atomic force calculation, structural stability comparison, and interactive structure visualization, transforming from a static analytical tool into a dynamic, self-improving research platform (refer to the **Supplementary Information** for implementation details).

## Data Availability

The AI Agents developed in this study and the comprehensive MLP training dataset (containing approximately 90,000 atomic configurations with corresponding DFT energies and forces) are both hosted on the Digital Automation for Scientific Discovery platform (DigAuto) at https://www.digauto.org. Key computational code, including the MCP module for the closed-loop AI agent framework, trained MLP models, and analysis scripts, is available in the supplementary information. Additional information is available upon reasonable request to the corresponding authors.

## Contributions

Y.M. Lu conceived and designed the study, developed the closed-loop AI agent framework and machine learning potential, performed all computational simulations and data analysis, and wrote the original manuscript. T.Y. Lu contributed to the training of machine learning potential. D. Zhang contributed to the conceptualization and provided key insights for the active learning approach. L.L. Ye supervised the overall research direction and provided strategic guidance. H. Li co-supervised the project and provided theoretical oversight. All authors discussed the results, reviewed the manuscript, and approved the final version.

# Acknowledgements

This work was supported by JSPS KAKENHI (Grant Nos. JP25H01508, JP25K17991, and JP24K23068), and the National Natural Science Foundation of China (No. 22309109). The authors acknowledge the Center for Computational Materials Science, Institute for Materials Research, Tohoku University for the use of MASAMUNE-IMR (Project Nos. 202412-SCKXX-0211 and 202412-SCKXX-0209), and the Institute for Solid State Physics (ISSP) at the University of Tokyo for the use of their supercomputing facilities.